\documentclass[runningheads]{llncs}

\usepackage[T1]{fontenc}
\usepackage{amsmath,amssymb,amsfonts}
\usepackage{graphicx}
\usepackage{subcaption}
\usepackage{booktabs}
\usepackage{multirow}
\usepackage{placeins}
\usepackage{float}
\usepackage{caption}
\usepackage{xcolor,colortbl}
\usepackage{textcomp}
\usepackage{tablefootnote}
\usepackage{pifont}

\makeatletter
\renewcommand{\section}{\@startsection{section}{1}{0pt}{9pt}{9pt}{\bfseries}}
\renewcommand{\subsection}{\@startsection{subsection}{2}{0pt}{8pt}{8pt}{\bfseries}}
\makeatother
\usepackage[linesnumbered,ruled,vlined]{algorithm2e}

\usepackage{hyperref}
\hypersetup{
    colorlinks=true,
    citecolor=green
}
\usepackage{cite}
\usepackage{orcidlink}

\begin{document}
\title{R\&D: Balancing Reliability and Diversity \\
in Synthetic Data Augmentation\\ for Semantic Segmentation}

\author{Quang-Huy Che\inst{1,2,\orcidlink{0009-0007-7477-4702}}, Dinh-Duy Phan\inst{1,2,}\thanks{Corresponding author}, Duc-Khai Lam\inst{1,2,\orcidlink{0000-0003-2711-1408}}}
\authorrunning{Quang-Huy Che et al.}
\institute{University of Information Technology, Ho Chi Minh City, Vietnam\\
\and
Vietnam National University, Ho Chi Minh City, Vietnam\\
\email{huycq@uit.edu.vn, duypd@uit.edu.vn, khaild@uit.edu.vn}}

\titlerunning{R\&D: Balancing Reliability and Diversity}

\maketitle
\begin{abstract}
Collecting and annotating datasets for pixel-level semantic segmentation tasks are highly labor-intensive. Data augmentation provides a viable solution by enhancing model generalization without additional real-world data collection. Traditional augmentation techniques, such as translation, scaling, and color transformations, create geometric variations but fail to generate new structures. While generative models have been employed to extend semantic information of datasets, they often struggle to maintain consistency between the original and generated images, particularly for pixel-level tasks. In this work, we propose a novel synthetic data augmentation pipeline that integrates controllable diffusion models. Our approach balances diversity and reliability data, effectively bridging the gap between synthetic and real data. We utilize \textit{class-aware prompting} and \textit{visual prior blending} to improve image quality further, ensuring precise alignment with segmentation labels. By evaluating benchmark datasets such as PASCAL VOC and BDD100K, we demonstrate that our method significantly enhances semantic segmentation performance, especially in data-scarce scenarios, while improving model robustness in real-world applications. Our code is available at \href{https://github.com/chequanghuy/Enhanced-Generative-Data-Augmentation-for-Semantic-Segmentation-via-Stronger-Guidance}{https://github.com/chequanghuy/Enhanced-Generative-Data-Augmentation-for-Semantic-Segmentation-via-Stronger-Guidance}.

\end{abstract}
\section{Introduction}

Deep learning has transformed the field of computer vision, where model performance depends not only on methodological advancements but also significantly on the quality and quantity of training data. Large-scale datasets, such as SA-1B \cite{sam}, and Imagenet \cite{imagenet}, have played a crucial role in driving progress across various computer vision tasks. However, collecting and annotating these datasets is labor-intensive, especially for complex and privacy-sensitive data. This challenge is particularly notable in semantic segmentation, where each pixel in an image must be accurately classified. While widely used datasets like PASCAL VOC \cite{voc}, BDD100K \cite{bdd100k} provide a strong foundation for training segmentation models, expanding or creating new datasets of similar scale remains a significant bottleneck. Consequently, data augmentation has emerged as a critical approach to enhancing model generalization without requiring additional real-world data collection. This technique not only increases data diversity but also reduces annotation costs, offering an efficient alternative for addressing challenges in semantic segmentation.

Traditional data augmentation methods such as rotation, scaling, flipping, or pixel-level manipulations (e.g., blurring, adjusting brightness, and contrast) enhance model accuracy by introducing geometric and color variations. However, these transformations do not generate new structural components, perspectives, or textures, thus limiting their ability to expand dataset diversity. More advanced techniques include partial image removal methods (e.g. Random Erasing \cite{RandomErasing}, Cutout), or image mixing techniques (e.g., Mosaic \cite{Mosaic}, Mixup \cite{Mixup}). However, most of these methods primarily expand the visual representation space without introducing new semantic information, thereby reducing their effectiveness in improving the model’s generalization capability.

Unlike previous data augmentation methods \cite{datasetdm}, generative models are trained directly on the target dataset to produce additional samples. However, since these models learn from the same data domain, the generated samples often lack diversity compared to the original data. Without fine-tuning the target dataset, the synthesized images tend to follow the distribution of the pre-trained model, not the desired distribution for data augmentation. Although generative models \cite{dataset_diff,diffumask,stronger0} can generate semantically diverse images, ensuring distributional alignment between the original and generated data remains a challenge. Additionally, semantic segmentation requires that generated samples preserve precise object shapes and structures, unlike classification \cite{class2,class3} or object detection tasks \cite{obj2}. To address these limitations, we propose a synthetic data augmentation pipeline for semantic segmentation based on generative models. In summary, the contributions of our work are as follows:

\begin{itemize}
    \item We propose a novel synthetic data augment pipeline that integrates two controllable diffusion models to generate synthetic datasets for semantic segmentation. This approach bridges the gap between synthetic datasets and real datasets, ensuring both the diversity and reliability of synthetic images.

    \item We integrate the proposed pipeline with the \textit{class-aware prompting} method we propose and \textit{visual prior blending} \cite{stronger0}. These combined methods enhance the quality of the generated images by ensuring that all relevant objects are included in the generated images and improving the alignment of the generated images with segmentation labels, thereby ensuring high accuracy and reliability in the synthetic datasets.

    \item We demonstrate the effectiveness of our approach through extensive experiments on standard benchmarks, including PASCAL VOC \cite{voc} and BDD100K \cite{bdd100k}. Our method consistently improves semantic segmentation performance, particularly in data-scarce scenarios.
    
\end{itemize}

\section{Related work}

\subsection{Image Generation}

Image generation is a significant research direction in computer vision and artificial intelligence, especially with the rapid advancement of deep learning models in recent years. Generative Adversarial Networks (GANs) \cite{gan}, as foundational models in image synthesis, have been widely used to generate high-resolution images. However, GANs often face optimization challenges, making it difficult for the model to fully capture the underlying data distribution. Recently, diffusion models (DMs) have emerged as a more advanced approach to image generation, enabling the model to approximate the data distribution more stably compared to GANs. Stable Diffusion (SD) \cite{sd,sdxl} is a variant of diffusion models that leverages the latent space instead of directly processing images in the pixel space. Through the cross-attention mechanism, SD can generate images based on various input modalities such as text, bounding boxes, or semantic maps. One key advancement that enhances controllability in image generation is the integration of SD with ControlNet \cite{controlnet} or T2I-Adapter \cite{t2iadapter}. These methods allow the model to incorporate additional structured guidances (visual priors), represented as edges, segmentation masks, lineart, and depth maps, improving the consistency of shape and structure in the generated images.

\subsection{Image Synthesis for Data Augmentation}

Previous studies have utilized Generative Adversarial Networks (GANs) to generate synthetic data for semantic segmentation, focusing primarily on object-centered images. However, these methods face limitations when handling complex image layouts or interactions between multiple objects. With advancements in generative models, data augmentation techniques based on diffusion models have recently emerged. However, most of these methods are tailored for image classification \cite{class2,class3} or object detection \cite{obj2} rather than semantic segmentation, which requires pixel-level precision. Semantic segmentation poses a significant challenge for generative image synthesis due to its strict accuracy requirements. \cite{dataset_diff,diffumask} introduced Synthetic Dataset approaches capable of generating synthetic images along with segmentation masks for specified classes. These methods generate synthetic datasets and pseudo-labels from text descriptions, enabling data utilization for pretraining segmentation models. Unlike synthetic dataset approaches, generative-based augmentation uses existing images and masks to create additional training data. Inpainting-based methods \cite{inpaint2} modify objects while preserving backgrounds but often limit data diversity. In contrast, Che et al. \cite{stronger0} introduced the Controllable Diffusion Model with strong guidance for image synthesis, demonstrating notable improvements in data augmentation. However, synthetic data generation faces two challenges: (1) mismatches between segmentation masks and synthesized images and (2) domain shifts due to the generative model's training dataset constraints.

In this work, we propose a synthetic data augmentation pipeline based on generative models. Our pipeline integrates advanced techniques to enhance the robustness of the generated data. Additionally, our method achieves a balance between diversity and data reliability consistency compared to the original dataset, resulting in high-quality synthetic data suitable for training semantic segmentation models.

\begin{figure}[!b]
    \centering
    \includegraphics[width=\linewidth]{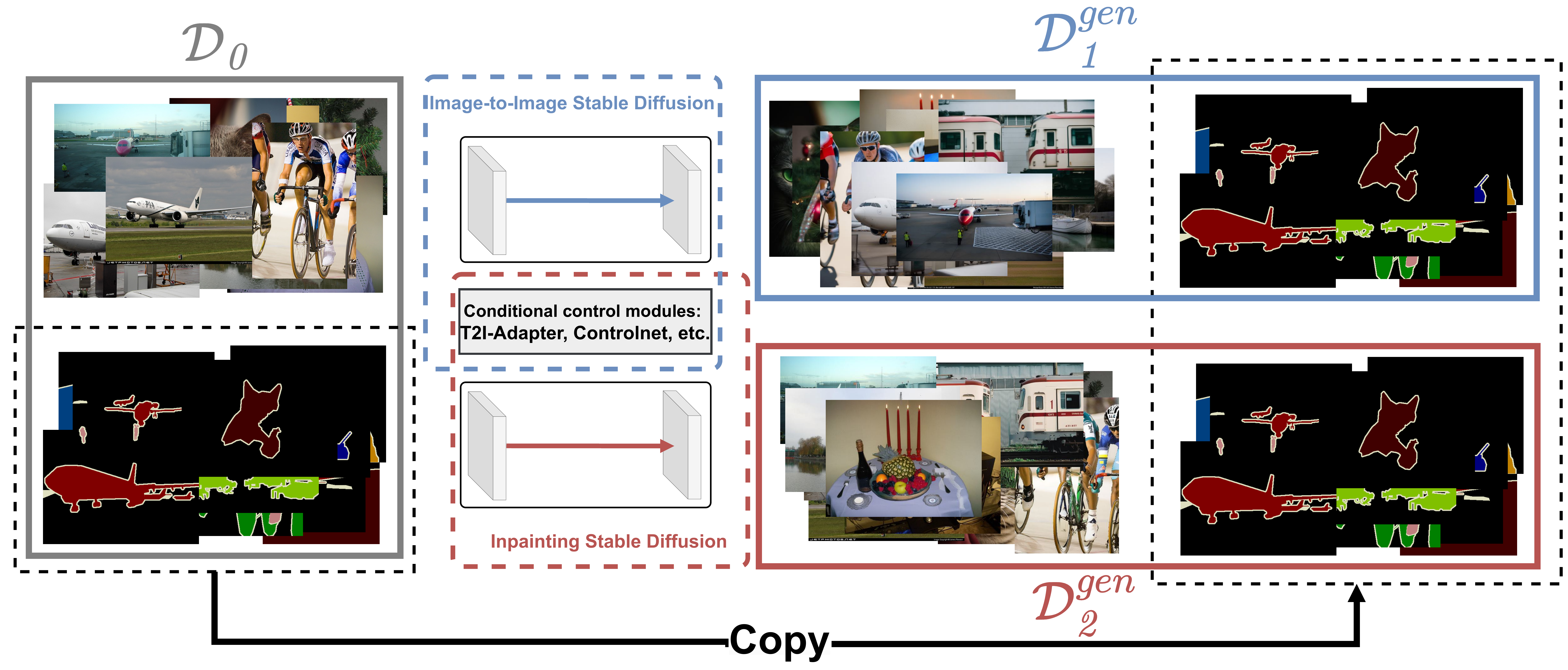}
    \caption{Our proposed synthetic data augmentation pipeline utilizes the real dataset \(\mathcal{D}_0\) to create two synthetic datasets, \(\mathcal{D}_1^{gen}\) and \(\mathcal{D}_2^{gen}\). The annotations for the synthetic data are directly copied from the labels of the real dataset.}
    \label{fig:pipeline}
\end{figure}

\section{Methods}

In this work, we propose a pipeline that integrates two SD models designed for controllable synthetic data generation: the Image-to-Image Controllable Diffusion Model (Sec. \ref{sec:control}) and the Controllable Inpainting Diffusion Model (Sec. \ref{sec:inpaint}). This pipeline takes an image and its corresponding segmentation labels as input and generates two synthetic images for each input image. The overall architecture of the proposed pipeline is illustrated in Fig. \ref{fig:pipeline}. Given a real dataset \(\mathcal{D}_0\), proposed pipeline generates the synthetic dataset \(\mathcal{D}_1^{gen} \cup \mathcal{D}_2^{gen}\), where \(\mathcal{D}_1^{gen}\) is generated by the Image-to-Image Controllable Diffusion Model, producing a highly diverse dataset by changing both labeled and unlabeled objects as well as the background, while \(\mathcal{D}_2^{gen}\) is generated by the Controllable Inpainting Diffusion Model, ensuring data distribution consistency by modifying only the labeled objects while keeping the remaining parts unchanged. The merging of the two datasets \(\mathcal{D}_1^{gen}\) and \(\mathcal{D}_2^{gen}\) yields a reliable synthetic dataset that simultaneously maximizes diversity and preserves data distribution fidelity. Furthermore, to enhance synthetic image precision, we propose novel methods for textual prompt refinement and visual prior in Section \ref{sec:condition}.

% In this work, we propose a pipeline that integrates two Stable Diffusion models designed for controllable generation: Controllable-Inpainting Diffusion Model and Image-to-Image Controllable Diffusion Model. The pipeline takes images and their corresponding segmentation labels as input and produces two generated images per input image, corresponding to the outputs of both diffusion models. The overall architecture of our proposed pipeline is illustrated in Fig. \ref{fig:pipeline}. 

\subsection{Robust condition for Diffusion Controllable Models} \label{sec:condition}

\subsubsection{Preparing text prompt:} 

To generate high-quality synthetic images for semantic segmentation, constructing an effective prompt is crucial in ensuring the presence of all relevant objects in the generated image. A straightforward approach is to list the annotated classes explicitly. Given an image $\mathcal{I}_i$ with list of labeled classes $\mathcal{C}_i = [c_1,$ $c_2,$ …], a simple prompt can be formulated as ``\texttt{A photograph of \texttt{c}$_1$, \texttt{c}$_2$,...}''. While this approach ensures that all objects in the image are mentioned, it lacks contextual information, making it challenging for the generative model to produce a coherent and realistic image. Instead of using simple annotated class lists, another approach is applying image captioning models to generate descriptions for datasets. However, these captions do not guarantee the inclusion of all annotated classes in the image, which may result in generated images that are either incomplete or contain incorrect objects.

\begin{figure}[t]
    \centering
    \includegraphics[width=\linewidth]{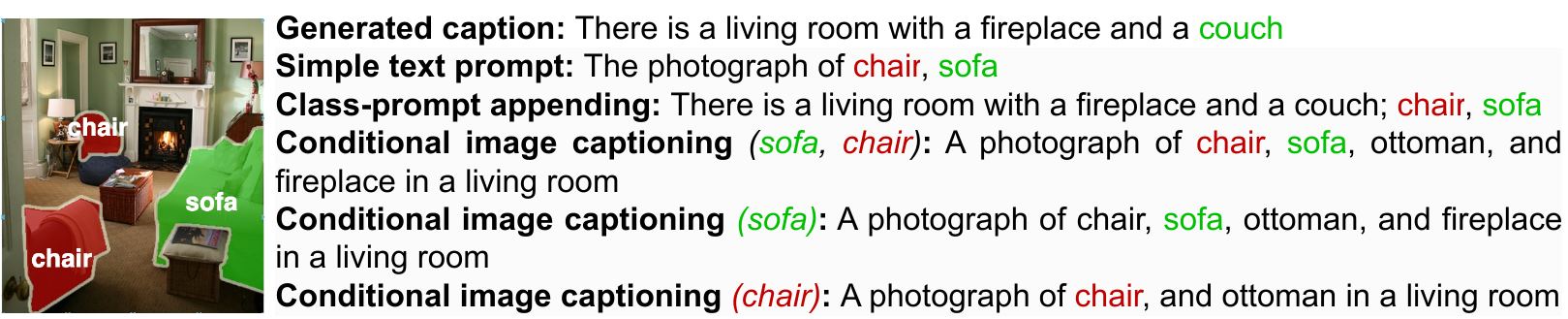}
    \caption{Some examples of text prompt selection for input images show that simple text prompts are often too simplistic, while generated captions may miss some labeled classes. Class-prompt appending addresses this but can lead to incoherent prompts. In contrast, conditional image captioning creates coherent prompts that accurately describe the image and include all labeled classes.}
    \label{fig:prompt}
\end{figure}

To address these challenges, we propose a prompt formulation integrating general contextual information about the image and a list of annotated classes. Unlike previous works \cite{dataset_diff,stronger0}, which merely concatenate the image caption with the annotated class list—often leading to poor linguistic coherence—we utilize BLIP \cite{blip} as an conditional image captioning. Specifically, BLIP generating a more comprehensive description that combines visual context with class labeled list. To enhance the focus on class tokens corresponding to target objects, we propose re-weighting mechanism during class token embedding \cite{compel}. By assigning higher weights to class tokens, this approach emphasizes key objects, thereby improving segmentation accuracy. The adjusted class tokens are denoted as ``[\textit{class}]++''. Our proposed prompt generation method, called \textit{class-aware prompting}, focuses on integrating class-specific information to produce more contextually rich and accurate prompts. Figure \ref{fig:prompt} illustrates an example of an image alongside its corresponding label, as well as various types of prompts.

\subsubsection{Visual priors for controllable model:}

Controllable generative models are characterized by their ability to generate high-quality images guided by visual priors. Among these, edge-based visual priors are widely used for object representation because they can generalize image structures. However, relying solely on edge information may introduce limitations when target objects are not well-emphasized or edge maps lack sufficient detail. This limitation can lead to generated objects not aligning accurately with the segmentation labels. To address this issue, our proposed pipeline incorporates \textit{visual prior blending} \cite{stronger0}, a technique designed to enhance the representation of labeled objects, ensuring that generated content better aligns with segmentation labels. Given ${V}^{I}$ as the visual prior derived from the original image and ${V}^{S}$ as the visual prior extracted from the segmentation mask, the blended visual prior ${V}^*$ is formulated as:

\begin{equation}
    {V}^* = \alpha {V}^{I} + {V}^{S}
\end{equation}
where $\alpha$ $\in$ (0, 1)  is a blending coefficient. Setting $\alpha$ < 1 reduces the influence of global image structures while emphasizing the information from labeled objects.

% Some examples of visual information blending are presented in Fig. \ref{fig:blending}.

% \begin{figure}
%     \centering
%     \includegraphics[width=\linewidth]{figures/blending.pdf}
%     \caption{Example of using the \textit{visual prior blending} method to enhance the visibility of labeled objects.}
%     \label{fig:blending}
% \end{figure}

% \subsection{Generation Pipeline}

\subsection{Image-to-Image Controllable Diffusion Model} \label{sec:control}

Controllable Diffusion Models \cite{t2iadapter,controlnet} have demonstrated remarkable capabilities in generating highly diverse synthetic images \cite{stronger0,obj2}, offering significant advantages for data augmentation techniques. The image generation can be process as a function \textbf{G}$_0$ : ${V}$ $\times$ ${P}$ $\to$ ${I}^{gen}$, where ${V}$ represents the visual prior of input image, ${P}$ denotes the textual prompt describing the image content, and ${I}^{gen}$ is the output image. However, controllable models often overlook input image distributions due to gaps between training data and target domains, leading to distribution shifts in generated images. To address this, we integrate an Image-to-Image (Img2Img) mechanism into the Controllable Diffusion Model framework. This extends the function \textbf{G}$_0$ to \textbf{G}$_1$ : ${I}$ $\times$ ${V}$ $\times$ ${P}$ $\to$ ${I}^{gen}$, where the additional input ${I}$ represents the reference image.

\begin{figure}[h]
    \centering
    \begin{subfigure}[b]{1.\textwidth}
        \centering
        \includegraphics[width=\textwidth]{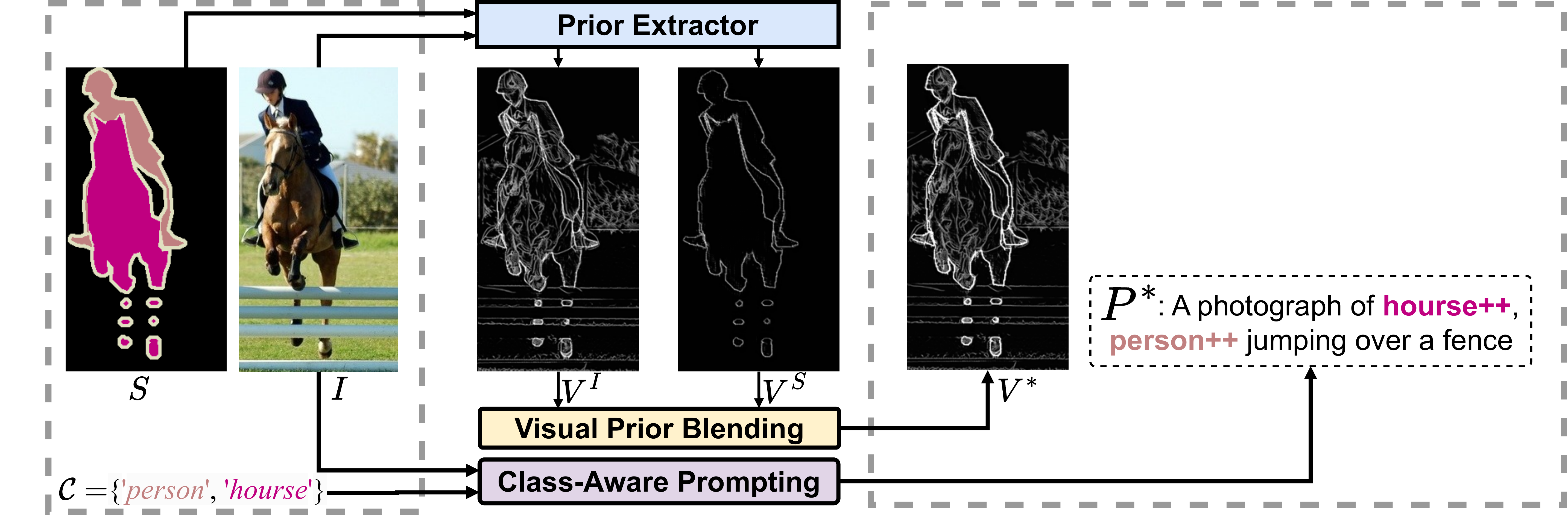}
        % \caption{The process of generating the stronger visual prior (${V}^*$) through the \textit{visual prior blending} method and the text prompt (${P}^*$) through \textit{class-aware prompting}.}
        \caption{}
        \label{fig:controla}
    \end{subfigure}

    \begin{subfigure}[b]{0.9\textwidth}
        \centering
        \includegraphics[width=\textwidth]{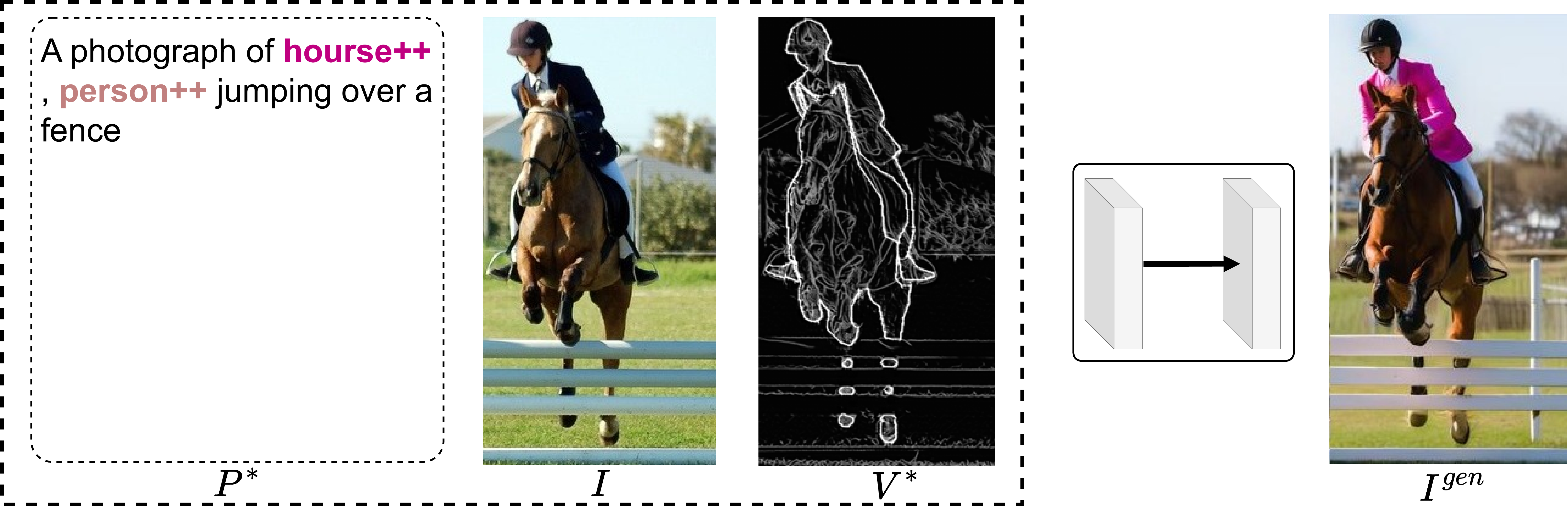}
        % \caption{The image generation process employs the Img2Img Controllable Diffusion Model, where the input image (${I}$) is combined with a text prompt (${P}^*$) and stronger visual prior (${V}^*$) to generate a diverse output image (${I}^{gen}$).}
        \caption{}
        \label{fig:controlb}
    \end{subfigure}

    \caption{Image generation using the Img2Img Controllable Diffusion Model.}
    \label{fig:main}
\end{figure}

This approach ensures that the generated images not only maintain diversity at a moderate level but also exhibit improved similarity to the reference image, thereby achieving better alignment with the target data distribution. Furthermore, the Img2Img mechanism preserves the reference image's structural composition more effectively than traditional Text-to-Image (T2I) methods, as it utilizes the input image as a foundational guide during the diffusion process. As illustrated in Fig. \ref{fig:controla}, our proposed pipeline incorporates two proposed methods to generate new image ${I}^{gen}$: \textit{class-aware prompting} and \textit{visual prior blending}, which generate ${P}^*$ and ${V}^*$, respectively. These components are then combined with input image, enabling the Img2Img Controllable Diffusion Model to produce highly diverse images while preserving fine-grained details and maintaining the distributional characteristics of the original data. 

\begin{equation}
{I}^{gen} = \textbf{G}^*_1( {I} , {V}^* , {P^*}) 
\end{equation}

The overall process is depicted in Fig. \ref{fig:controlb}, highlighting the model's ability to produce highly diverse images while preserving fine-grained details and maintaining the distributional characteristics of the original data.

\subsection{Controllable-Inpainting Diffusion Model}
\label{sec:inpaint}

\begin{figure}[t]
    \centering
    \begin{subfigure}[b]{1.\textwidth}
        \centering
        \includegraphics[width=1\textwidth]{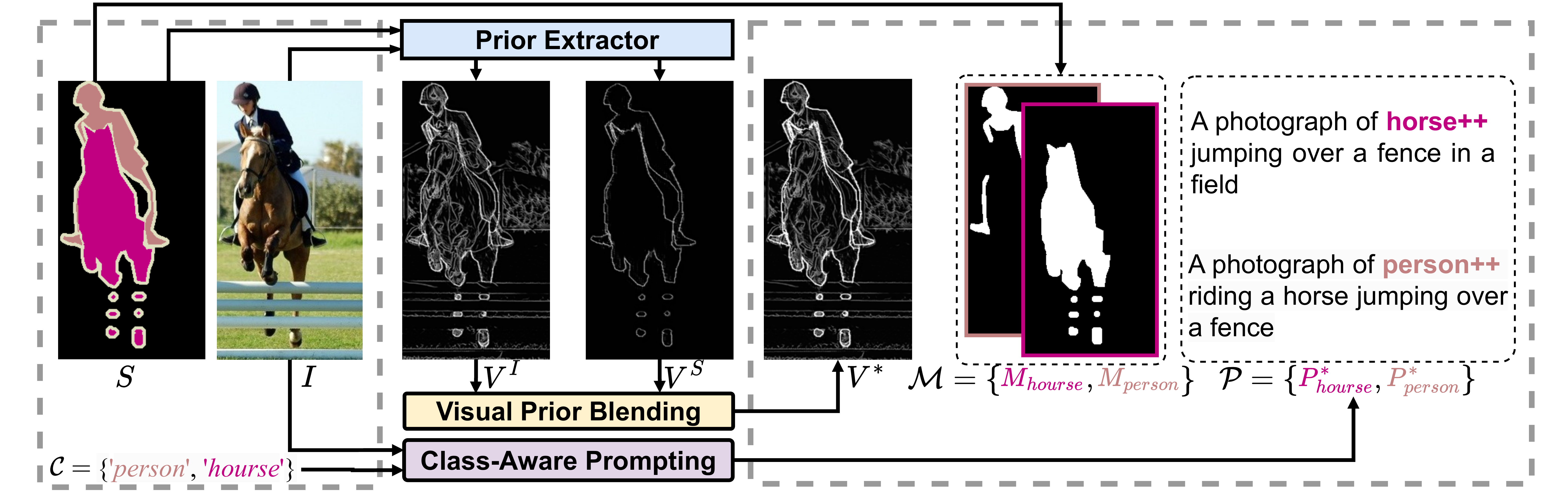}
        \caption{}
        \label{fig:inpainta}
    \end{subfigure}

    \begin{subfigure}[b]{1,0\textwidth}
        \centering
        \includegraphics[width=\textwidth]{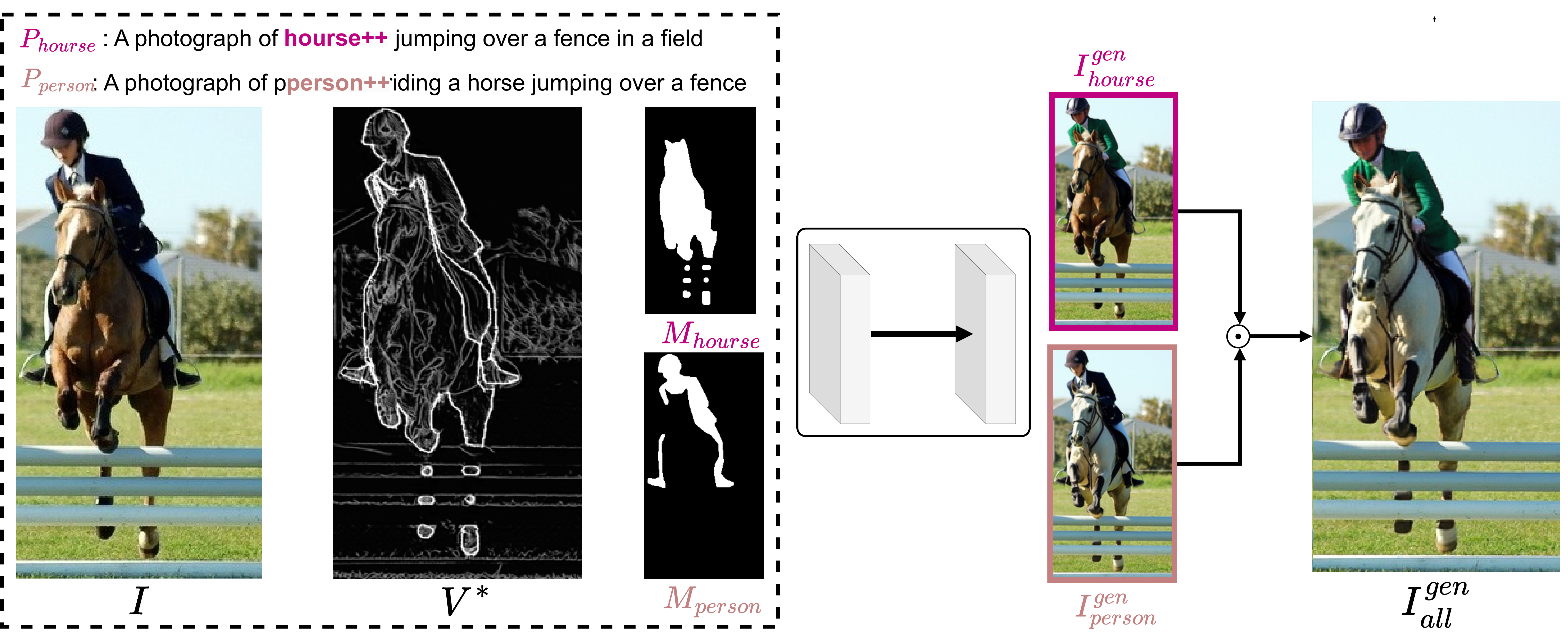}
        \caption{}
        \label{fig:inpaintb}
    \end{subfigure}

    \caption{Image generation using the Controllable Inpainting Diffusion Model.}
    \label{fig:img2img}
\end{figure}

The issue of data generation out of the original domain when using controllable diffusion models has been highlighted in previous research \cite{stronger0}. This phenomenon can lead to a decline in model training performance as the dataset size increases. Although Sec. \ref{sec:control} introduces the Image-to-Image Controllable Diffusion Model to mitigate this limitation, the transformation of the entire image makes it challenging to preserve the original data distribution. Therefore, we proposed to maintain the original image characteristics by employing an Inpainting Diffusion Model to modify specific regions of the image instead of transforming the entire image. To balance the advantages of both methods for the data augmentation task, we combine the Inpainting Diffusion Model with the proposed Img2Img Controllable Diffusion Model in Sec. \ref{sec:control}, aiming to balance data diversity and the reliability of the generated images. The transformation function can represent the process of the Inpainting Diffusion model \textbf{G} : ${I}$ $\times$ ${M}$ $\times$ ${P}$ $\to$ ${I}^{gen}$. Here, ${I}$ is the input image, ${M}$ is the mask specifying the regions to be modified, ${P}$ represents the visual prior controlling the inpainting process, and ${I}^{gen}$ is the generated image after the inpainting process.

\cite{inpaint2} has been noted that relying on a pre-existing Inpainting Diffusion Model \textbf{G} does not ensure the newly generated objects conform to the mask ${M}$. Additionally, it does not guarantee that the generated objects retain the original objects. To address this limitation, we integrate the Controllable Model with the Inpainting Diffusion Model, resulting in a novel framework termed the Controllable Inpainting Diffusion Model. Precisely, controllable models such as T2I-Adapter \cite{t2iadapter} and ControlNet \cite{controlnet} extract visually structured information to inject into the Unet architecture of the Diffusion Model. In this approach, the generation process is conditioned not only on the mask ${M}$ but also on the visual prior information ${V}$. We also utilize \textit{class-aware prompting} and \textit{visual prior blending} techniques to reduce the risk of small objects being removed and replaced with background elements or inaccurately generated shapes; this process is illustrated in Fig. \ref{fig:inpainta}. These improvements tackle the limitations associated to extends the function \textbf{G}$^*_2$: ${I}$ $\times$ ${V}^*$ $\times$ ${M}$ $\times$ ${P}^*$ $\to$ ${I}^{gen}$. However, unlike the approach using the Img2Img Controllable Diffusion Model, the generation of objects in the Controllable Inpainting Diffusion Model is performed sequentially for each object type. This approach allows for a more accurate generation of objects, particularly when dealing with objects that have similar shapes. In this case, the images for each new object are generated as follows:

\begin{equation}
I_i^{gen} = \textbf{G}^*_2( {I} , {V}^* , {M}_i , {P}^*_i) 
\end{equation}
where ${M}_i \in \mathcal{M}$ is the segmentation mask for the $i$-th class ($c_i \in \mathcal{C}$), determined by leveraging segmentation labels for each class. Meanwhile, ${P}_i^* \in \mathcal{P}^* $ is the prompt describing the data generation process for the class $c_i$. After obtaining the list of images \( \{{I}_i\} \) generated based on the information for each class, we perform an image merging operation to produce the final composite image \( {I}_{all}^{gen} \). This process relies on the list of masks \( \{M_i\} \) to obtain an image with the labeled objects modified. This process is shown in Fig. \ref{fig:inpaintb}.

\begin{equation}
    {I}^{gen}_{all} =  {I} \odot \Bigl(1-\sum_{i=1}^N{M_i} \Bigl)+\sum_{i=1}^N{\Bigl(I^{gen}_i \odot M_i\Bigl)}
\end{equation}

\begin{table*}[!b]
\caption{Comparison of mIoU (\%) on the validation set when training models on the original dataset ($\mathcal{D}_0$) and when merging it with the synthetic dataset ($\mathcal{D}_0$ $\cup$ \cite{stronger0} vs. $\mathcal{D}_0$ $\cup$ ours), using DeepLabV3+ and Mask2Former model architectures.}
\label{tab:main_table}
% \resizebox{\textwidth}{!}{%
\centering
\begin{tabular}{lllcllcccccccccccc}
\toprule
\textbf{Dataset}             &                            &  &                                    & \textbf{} &  & \textbf{VOC7}  & \multicolumn{1}{l}{\textbf{}} & \multicolumn{1}{l}{} & \multicolumn{9}{c}{\textbf{VOC12}}                                                                                                                                             \\ \hline
Number images                &                            &  &                                    &           &  & 209            & \multicolumn{1}{l}{}          & \multicolumn{1}{l}{} & 92             & \multicolumn{1}{l}{} & 183            & \multicolumn{1}{l}{} & 366            & \multicolumn{1}{l}{} & 732            & \multicolumn{1}{l}{} & 1464           \\ \midrule
\multirow{6}{*}{DeepLabV3+}  & \multirow{3}{*}{Resnet50}  &  & $\mathcal{D}_0$                       &           &  & 63.75          &                               &                      & 48.19          &                      & 58.44          &                      & 65.84          &                      & \underline{70.55} & \underline{}            & \underline{72.19} \\ \cmidrule(lr){4-18} 
                             &                            &  & $\mathcal{D}_0$ $\cup$ \cite{stronger0} &           &  & \underline{64.02} & \underline{}                     & \underline{}            & \underline{51.83} & \underline{}            & \underline{59.37} & \underline{}            & \underline{65.98} &                      & 69.14          &                      & 72.16          \\ \cmidrule(lr){4-18} 
                             &                            &  & $\mathcal{D}_0$ $\cup$ ours             &           &  & \textbf{64.47} & \textbf{}                     & \textbf{}            & \textbf{53.67} & \textbf{}            & \textbf{59.98} & \textbf{}            & \textbf{67.52} & \textbf{}            & \textbf{71.06} & \textbf{}            & \textbf{72.96} \\ \cmidrule(lr){2-18} 
                             & \multirow{3}{*}{Resnet101} &  & $\mathcal{D}_0$                      &           &  & 67.61          &                               &                      & 54.06          &                      & 62.88          &                      & 67.85          &                      & \underline{73.06} & \underline{}            & \underline{76.19} \\ \cmidrule(lr){4-18} 
                             &                            &  & $\mathcal{D}_0$ $\cup$  \cite{stronger0} &           &  & \underline{68.79} &                               &                      & \textbf{56.01} &                      & \underline{63.09} & \underline{}            & \underline{68.89} &                      & 73.05          &                      & 75.68          \\ \cmidrule(lr){4-18} 
                             &                            &  & $\mathcal{D}_0$ $\cup$ ours             &           &  & \textbf{68.81} &                               &                      & \underline{55.93} &                      & \textbf{64.08} & \textbf{}            & \textbf{69.54} & \textbf{}            & \textbf{73.83} & \textbf{}            & \textbf{76.91} \\ \midrule
\multirow{3}{*}{Mask2Former} & \multirow{3}{*}{Swin-B}    &  & $\mathcal{D}_0$                       &           &  & 76.19          &                               &                      & 59.11          &                      & 74.39          &                      & 75.21          &                      & 79.02          &                      & 81.78          \\ \cmidrule(lr){4-18} 
                             &                            &  & $\mathcal{D}_0$ $\cup$  \cite{stronger0} &           &  & \underline{77.01} &                               &                      & \textbf{65.01} & \textbf{}            & \textbf{76.67} &                      & \underline{77.10} & \underline{}            & \underline{79.87} & \underline{}            & \underline{81.86} \\ \cmidrule(lr){4-18} 
                             &                            &  & $\mathcal{D}_0$ $\cup$ ours             &           &  & \textbf{78.52} &                               &                      & \underline{64.51} & \underline{}            & \underline{76.45} &                      & \textbf{77.84} & \textbf{}            & \textbf{81.08} & \textbf{}            & \textbf{82.88} \\ \bottomrule
\end{tabular}%
\end{table*}

\section{Experiments}

\subsection{Datesets and implementation details}

\subsubsection{Datesets:}We evaluate our synthetic data generation framework on two benchmark datasets: PASCAL VOC \cite{voc} (VOC07 and VOC12) and BDD100K \cite{bdd100k}. To assess performance under data-limited scenarios, we conduct experiments on both the full VOC12 dataset (1,464 images), and its subsets \cite{subset}. Beyond standard object segmentation tasks, we further validate the model’s capability to generate images under diverse environmental conditions (e.g., weather, scene) using BDD100K, a large-scale dataset capturing real-world driving scenarios for drivable area and lane segmentation tasks.

\subsubsection{Implementation details:}For object segmentation evaluation, we employ Deep\\
-LabV3+ \cite{deeplabv3} (with ResNet50/101 backbones) and Mask2Former \cite{mask2former} (Swin-B backbone) implemented in the MMSegmentation framework, training for 30K iterations at 512×512 resolution with batch size 16 using AdamW optimization and default augmentations. For weather-conditioned segmentation on BDD100K \cite{bdd100k}, we adopt TwinLiteNet \cite{twin} for simultaneous lane and drivable area segmentation. Our image generation pipeline leverages SD-XL \cite{sdxl} controlled via T2I-Adapter \cite{t2iadapter} with Line Art as visual priors. The coefficient $\alpha$ in the \textit{visual prior blending} method is set to 0.8.

\begin{table}[t]
\caption{The table shows the performance of the Mask2Former (Swin-B) model when trained on (1) the real dataset, (2) the synthetic dataset, and (3) fine-tuned on the real dataset after pre-training on the synthetic dataset. The compared methods include generating synthetic data with pseudo-labels \cite{diffumask,dataset_diff,attn2mask} and synthetic data based on the original dataset \cite{stronger0}}
\label{tab:tune}
\centering
\resizebox{1.\textwidth}{!}{
\begin{tabular}{lcclcccccllc}
\toprule
                      & \multicolumn{2}{c}{\textbf{Real images}}                                                                                           & \multicolumn{1}{c}{\textbf{}} & \multicolumn{5}{c}{\textbf{Synthetic images}}                                                                                                                                                                                                                                                                                                                                                                                    & \multicolumn{1}{c}{} & \multicolumn{1}{c}{} &                                      \\ \cmidrule(lr){2-3} \cmidrule(lr){5-9}
                      & \cellcolor[HTML]{EFEFEF}\begin{tabular}[c]{@{}c@{}}\textbf{VOC} \\ (5k)\end{tabular} & \begin{tabular}[c]{@{}c@{}}\textbf{VOC}\\ (1,5k)\end{tabular} & \multicolumn{1}{c}{}          & \cellcolor[HTML]{EFEFEF}\begin{tabular}[c]{@{}c@{}}\textbf{DiffuMask}\\ \cite{diffumask} (60k)\end{tabular} & \begin{tabular}[c]{@{}c@{}}\textbf{DD}\\ \cite{dataset_diff} (40k)\end{tabular} & \cellcolor[HTML]{EFEFEF}\begin{tabular}[c]{@{}c@{}}\textbf{Attn2Mask} \\ \cite{attn2mask}\end{tabular} & \begin{tabular}[c]{@{}c@{}}\textbf{SG}\\ \cite{stronger0} (1,5k)\end{tabular} & \cellcolor[HTML]{EFEFEF}\begin{tabular}[c]{@{}c@{}}\textbf{Ours}\\ (2,9k)\end{tabular} & \multicolumn{1}{c}{} & \multicolumn{1}{c}{} & \multirow{-2}{*}{\textbf{mIoU (\%)}} \\ \hline
                      & \cellcolor[HTML]{EFEFEF}$\checkmark$                                        &                                                      &                               & \cellcolor[HTML]{EFEFEF}                                                                           &                                                                        & \cellcolor[HTML]{EFEFEF}                                                                    &                                                                      & \cellcolor[HTML]{EFEFEF}                                                      &                      &                      & 83.4                                 \\
\multirow{-2}{*}{(1)} & \cellcolor[HTML]{EFEFEF}                                                    & $\checkmark$                                         &                               & \cellcolor[HTML]{EFEFEF}                                                                           &                                                                        & \cellcolor[HTML]{EFEFEF}                                                                    &                                                                      & \cellcolor[HTML]{EFEFEF}                                                      &                      &                      & 81.8                                 \\ \hline
                      & \cellcolor[HTML]{EFEFEF}                                                    &                                                      &                               & \cellcolor[HTML]{EFEFEF}$\checkmark$                                                               &                                                                        & \cellcolor[HTML]{EFEFEF}                                                                    &                                                                      & \cellcolor[HTML]{EFEFEF}                                                      &                      &                      & 70.6                                 \\
                      & \cellcolor[HTML]{EFEFEF}                                                    &                                                      &                               & \cellcolor[HTML]{EFEFEF}                                                                           & $\checkmark$                                                           & \cellcolor[HTML]{EFEFEF}                                                                    &                                                                      & \cellcolor[HTML]{EFEFEF}                                                      &                      &                      & 67.6                                 \\
                      & \cellcolor[HTML]{EFEFEF}                                                    &                                                      &                               & \cellcolor[HTML]{EFEFEF}                                                                           &                                                                        & \cellcolor[HTML]{EFEFEF}$\checkmark$                                                        &                                                                      & \cellcolor[HTML]{EFEFEF}                                                      &                      &                      & 71.0                                 \\
                      & \cellcolor[HTML]{EFEFEF}                                                    &                                                      &                               & \cellcolor[HTML]{EFEFEF}                                                                           &                                                                        & \cellcolor[HTML]{EFEFEF}                                                                    & $\checkmark$                                                         & \cellcolor[HTML]{EFEFEF}                                                      &                      &                      & 73.0                                 \\
\multirow{-5}{*}{(2)} & \cellcolor[HTML]{EFEFEF}                                                    &                                                      &                               & \cellcolor[HTML]{EFEFEF}                                                                           &                                                                        & \cellcolor[HTML]{EFEFEF}                                                                    &                                                                      & \cellcolor[HTML]{EFEFEF}$\checkmark$                                          &                      &                      & 76.3                                 \\ \hline
                      & \cellcolor[HTML]{EFEFEF}$\checkmark$                                        &                                                      &                               & \cellcolor[HTML]{EFEFEF}$\checkmark$                                                               &                                                                        & \cellcolor[HTML]{EFEFEF}                                                                    &                                                                      & \cellcolor[HTML]{EFEFEF}                                                      &                      &                      & 84.9                                 \\
                      & \cellcolor[HTML]{EFEFEF}                                                    & $\checkmark$                                         &                               & \cellcolor[HTML]{EFEFEF}                                                                           & $\checkmark$                                                           & \cellcolor[HTML]{EFEFEF}                                                                    &                                                                      & \cellcolor[HTML]{EFEFEF}                                                      &                      &                      & 82.4                                 \\
                      & \cellcolor[HTML]{EFEFEF}                                                    & $\checkmark$                                         &                               & \cellcolor[HTML]{EFEFEF}                                                                           &                                                                        & \cellcolor[HTML]{EFEFEF}                                                                    & $\checkmark$                                                         & \cellcolor[HTML]{EFEFEF}                                                      &                      &                      & 82.8                                 \\
\multirow{-4}{*}{(3)} & \cellcolor[HTML]{EFEFEF}                                                    & $\checkmark$                                         &                               & \cellcolor[HTML]{EFEFEF}                                                                           &                                                                        & \cellcolor[HTML]{EFEFEF}                                                                    &                                                                      & \cellcolor[HTML]{EFEFEF}$\checkmark$                                          &                      &                      & 84.0                                 \\ \bottomrule
\end{tabular}%
}
\end{table}

\begin{table}[!b]
\centering
\caption{Evaluation of multi-task segmentation model performance across different environmental conditions when integrating our method with real data via merging/fine-tuning.}
\label{twin}
\begin{tabular}{llclclclcllc}
\toprule
\multirow{2}{*}{\textbf{Condition}} &  & \multirow{2}{*}{\textbf{Ours}} &  & \multirow{2}{*}{\textbf{Number}} &  & \multicolumn{3}{c}{\textbf{Lane Line}}                      &                      &                      & \textbf{Drivable Area} \\ \cline{7-9} \cline{12-12} 
                                    &  &                                &  &                                  &  & \textbf{Accuracy (\%)} &                      & \textbf{IoU(\%)} &                      &                      & \textbf{mIoU(\%)}      \\ \hline
\multirow{2}{*}{Foggy}              &  &                                &  & 130                              &  & 56.9              &                      & 4.5              &                      &                      & 72.6                   \\ \cline{3-12} 
                                    &  & $\checkmark$                   &  & 390                              &  & \textbf{67.8} / 67.5              &                      & 8.3 / \textbf{8.7}              &                      &                      & 79.9 / \textbf{80.4}                \\ \hline
\multirow{2}{*}{Tunnel}             &  &                                &  & 129                              &  & 80.5              &                      & 9.3              &                      &                      & 73.4                   \\ \cline{3-12} 
                                    &  & $\checkmark$                   &  & 387                              &  & 84.4 / \textbf{85.2}              &                      & \textbf{15.2} / 14.9             &                      &                      & 86.0 / \textbf{87.3}                   \\ \hline
\multirow{2}{*}{Gas Station}        &  &                                &  & 27                               &  & 48.9              & \multicolumn{1}{c}{} & 0.2              & \multicolumn{1}{c}{} & \multicolumn{1}{c}{} & 67.1                   \\ \cline{3-12} 
                                    &  & $\checkmark$                   &  & 81                               &  & 62.6 / \textbf{63.7}              & \multicolumn{1}{c}{} & 0.4 / \textbf{0.5}               & \multicolumn{1}{c}{} & \multicolumn{1}{c}{} & 70.5 / \textbf{72.1}                  \\ \hline
\end{tabular}%
\end{table}

\subsection{Semantic segmentation result on VOC} 

To evaluate our proposed data augmentation method, we compare models (Deep-\\
LabV3+ and Mask2Former) trained on the original dataset ($\mathcal{D}_0$) and our augmented dataset ($\mathcal{D}_0 \cup \mathcal{D}_1^{gen} \cup \mathcal{D}_2^{gen}$). We also compare our method with Stronger Guidance \cite{stronger0}, re-implemented using our training settings. Notably, we did not apply the object filter \cite{obj2} or class balancing algorithm \cite{stronger0} to focus solely on synthetic image quality. As shown in Tab. \ref{tab:main_table}, our method consistently improves semantic segmentation performance across datasets and architectures. While our method outperforms the baseline (trained on $\mathcal{D}_0$) in all configurations, it occasionally underperforms compared to \cite{stronger0} on smaller datasets but not significantly (VOC12 with 92 images when trained on DeepLabV3 Resnet101 and VOC12 with 92 or 183 images when trained on Mask2Former). However, as the dataset size increases (e.g., VOC12 with 732 or 1464 images), our method consistently outperforms \cite{stronger0}, which sometimes underperforms the baseline. This suggests that while highly diverse data improves accuracy on small datasets, it may cause distribution shifts that decrease performance as the number of samples in the dataset grows. Our method addresses this by balancing diversity and data consistency.

Additionally, we follow \cite{diffumask,attn2mask}, first training on synthetic data and then fine-tuning on real data (VOC12 with 1464 images), as synthetic data may not perfectly align with real data or domain shifts. Tab. \ref{tab:tune} shows that our method achieves the highest mIoU (76.3\%) when trained solely on synthetic data, outperforming other approaches. After fine-tuning, DiffuMask \cite{diffumask} achieves the best performance (84.9\%), but requires 60k synthetic images for pre-training and 5k real images for fine-tuning. In contrast, our method achieves a competitive 84.0\% mIoU with only 2.9k synthetic images and 1.5k real images, improving the baseline (81.8\%) by 2.2\%. This highlights the effectiveness of our approach.

\subsection{Image generation based on environmental conditions}

% In addition to focusing on object segmentation, we also evaluate the ability to generate images based on different environmental conditions. In this evaluation, instead of ensuring the accurate generation of objects, we experiment with generating images under-emphasized environmental conditions, such as foggy, tunnel, and gas station scenarios. We train the TwinLiteNet model \cite{twin} for drivable area and lane segmentation tasks on the BDD100K dataset under different conditions. The results show that with limited data (fewer than 200 samples), the model performs poorly under different environmental conditions. However, the model’s performance improves significantly when our method is applied (through merging synthetic and real datasets or fine-tuning on real datasets). This highlights the capability of our method to generate synthetic data that aligns with specific environmental conditions, thereby enhancing the model’s performance in real-world scenarios.

In addition to object segmentation, we evaluate the generation of images under different environmental conditions, such as fog, tunnel, and gas station scenarios. Using the TwinLiteNet model \cite{twin} for the drivable area and lane segmentation on the BDD100K dataset, we observe poor performance with fewer than 200 samples in these conditions. However, applying our method—through merging synthetic and real datasets or fine-tuning on real data—significantly improves the model's performance. This highlights our method's ability to generate synthetic data tailored to specific environmental conditions, enhancing model performance in real-world scenarios.

\begin{figure}[t]
\centering
\begin{minipage}{0.3\textwidth}
\centering
\captionof{table}{Quantitative comparison (FID /CLIP Score (ViT-B/32)).}
\label{tab:fid}
\resizebox{\textwidth}{!}{ % Thay đổi kích thước bảng theo chiều rộng của trang
\begin{tabular}{lcc}
\toprule
                      & \textbf{CLIP $\uparrow$} & \textbf{FID $\downarrow$} \\ \midrule
\cite{stronger0}        & 0.81                     & 114.49                    \\ \midrule
\rowcolor[HTML]{EFEFEF} 
$\mathcal{D}^{gen}_1$ & 0.84                     & 101.92                    \\ \midrule
\rowcolor[HTML]{EFEFEF} 
$\mathcal{D}^{gen}_2$ & 0.92                     & 72.22                     \\ \bottomrule
\end{tabular}%
}
\end{minipage}
\hfill
\begin{minipage}{0.66\textwidth}
\centering
\includegraphics[width=1.\linewidth]{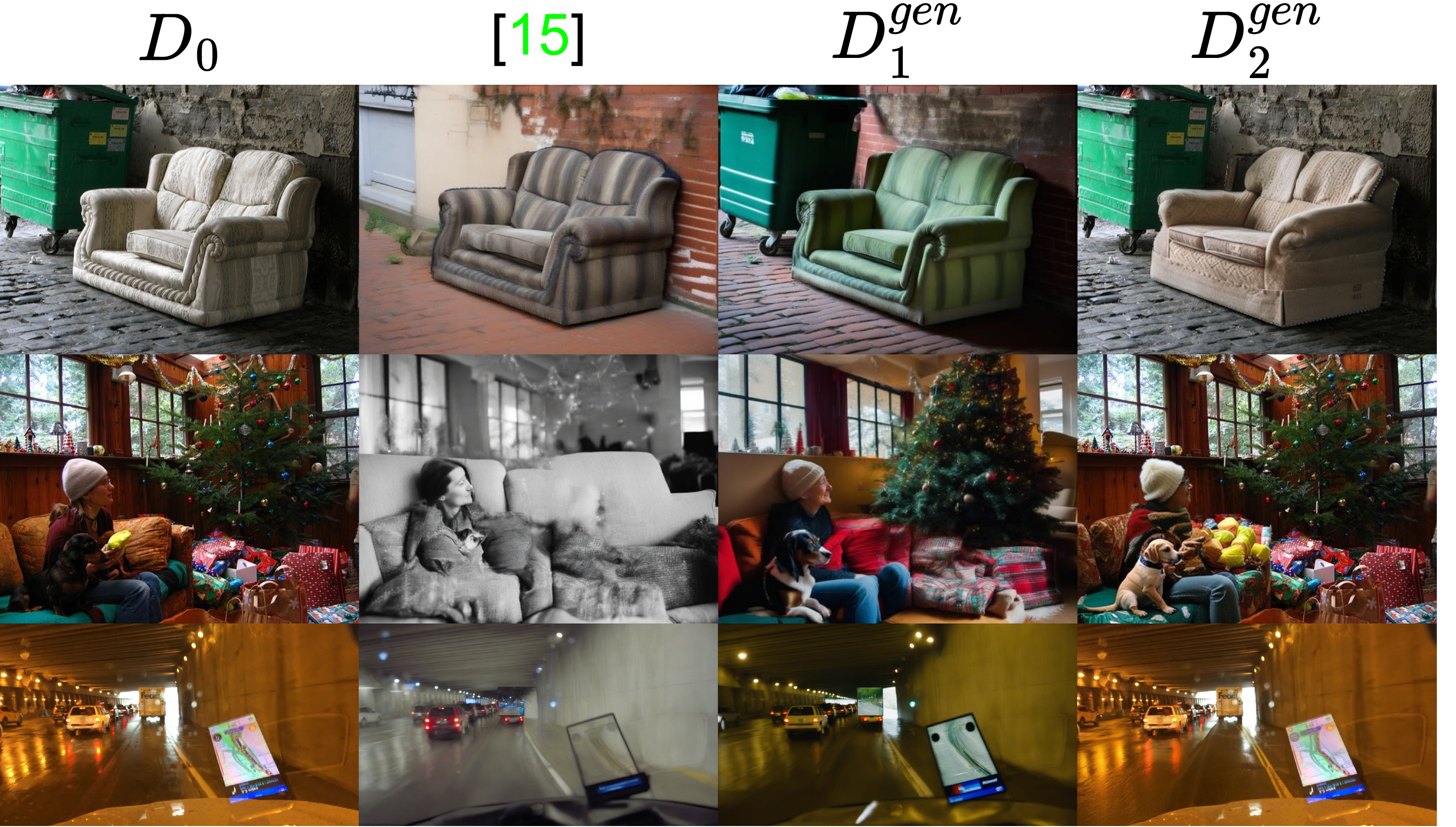} % Change to your actual figure path
\caption{Some synthetic images generated using different methods.}
\label{fig:visual}
\end{minipage}
\end{figure}

\subsection{Visualization and metrics for generated image quality}

Qualitative results on the PASCAL VOC and BDD100K datasets, as illustrated in Figure \ref{fig:visual}, demonstrate that our method generates highly similar images to the original images ($D_0$). Specifically, images produced by the Img2Img Controllable Diffusion model ($D^{gen}_1$) and the Controllable Inpainting Diffusion Model ($D^{gen}_2$) not only exhibit diversity but also maintain structural similarity to the original images. In contrast, the method proposed in \cite{stronger0} yields inferior results, failing to preserve the structure and distribution of the generated images. In addition, we conducted quantitative evaluations using two metrics: FID and CLIP Score. The results in Tab. \ref{tab:fid}, evaluated on the VOC7 dataset, show that our method achieves higher scores, confirming its ability to generate high-quality images that closely align with the distribution of the original data.

\subsection{Ablation Study}

This section presents a comprehensive ablation study evaluating our method's components using the PASCAL VOC with Mask2Former.

\begin{table}[t]
\centering
\caption{Comparison of semantic segmentation performance (mIoU\%) using different training strategies. Results are reported for Mask2Former models.}
\label{tab:ab_data}
% \resizebox{\columnwidth}{!}{%
\resizebox{1.\textwidth}{!}{
\begin{tabular}{lclclccccclclcclc}
\hline
                                                                                                             &                                      &                          &                                        &                          &                                        &                          &                          &                          & \multicolumn{3}{c}{\textbf{VOC7}}                                                                                                                                                          &                          & \textbf{}                & \multicolumn{3}{c}{\textbf{VOC12}}                                                                                                                                                           \\ \cmidrule(lr){10-12} \cmidrule(lr){15-17} 
\multirow{-2}{*}{}                                                                                           & \multirow{-2}{*}{\textbf{$\mathcal{D}_{0}$}}   &                          & \multirow{-2}{*}{\textbf{$\mathcal{D}^{gen}_1$}} &                          & \multirow{-2}{*}{\textbf{$\mathcal{D}^{gen}_2$}} &                          & \textbf{}                & \textbf{}                & \textbf{Number}                                                                 &                          & \textbf{mIoU (\%)}                                                            &                          & \textbf{}                & \textbf{Number}                                                                   &                          & \textbf{mIoU (\%)}                                                            \\ \hline
\begin{tabular}[c]{@{}l@{}}Train with Pure \\ Real Data\end{tabular}                                         & $\checkmark$                         &                          &                                        &                          &                                        &                          &                          &                          & R: 209                                                                          &                          & 76.19                                                                         &                          &                          & R: 1464                                                                           &                          & 81.8                                                                          \\ \hline
                                                                                                             &                                      &                          & $\checkmark$                           &                          &                                        &                          &                          &                          & S: 209                                                                          &                          & 74.32                                                                         &                          &                          & R: 1464                                                                           &                          & 74.22                                                                         \\
                                                                                                             &                                      &                          &                                        &                          & $\checkmark$                           & \multicolumn{1}{l}{}     & \multicolumn{1}{l}{}     & \multicolumn{1}{l}{}     & S: 209                                                                          &                          & 73.85                                                                         &                          &                          & R: 1464                                                                           &                          & 75.18                                                                         \\
\multirow{-3}{*}{\begin{tabular}[c]{@{}l@{}}Train with Pure \\ Synthetic Data\end{tabular}}                  &                                      &                          & $\checkmark$                           &                          & $\checkmark$                           &                          &                          &                          & S: 418                                                                          &                          & 74.67                                                                         &                          &                          & R: 2928                                                                           &                          & 76.27                                                                         \\ \hline
                                                                                                             & \cellcolor[HTML]{EFEFEF}$\checkmark$ & \cellcolor[HTML]{EFEFEF} & \cellcolor[HTML]{EFEFEF}$\checkmark$   & \cellcolor[HTML]{EFEFEF} & \cellcolor[HTML]{EFEFEF}               & \cellcolor[HTML]{EFEFEF} & \cellcolor[HTML]{EFEFEF} & \cellcolor[HTML]{EFEFEF} & \cellcolor[HTML]{EFEFEF}\begin{tabular}[c]{@{}c@{}}R: 209\\ S: 209\end{tabular} & \cellcolor[HTML]{EFEFEF} & \cellcolor[HTML]{EFEFEF}\begin{tabular}[c]{@{}c@{}}\textcolor{blue}{77.58}\\ \textcolor{red}{78.53}\end{tabular} & \cellcolor[HTML]{EFEFEF} & \cellcolor[HTML]{EFEFEF} & \cellcolor[HTML]{EFEFEF}\begin{tabular}[c]{@{}c@{}}R: 1464\\ S: 1464\end{tabular} & \cellcolor[HTML]{EFEFEF} & \cellcolor[HTML]{EFEFEF}\begin{tabular}[c]{@{}c@{}}\textcolor{blue}{81.91}\\ \textcolor{red}{83.01}\end{tabular} \\
                                                                                                             & $\checkmark$                         &                          &                                        &                          & $\checkmark$                           & \multicolumn{1}{l}{}     & \multicolumn{1}{l}{}     & \multicolumn{1}{l}{}     & \begin{tabular}[c]{@{}c@{}}R: 209\\ S: 209\end{tabular}                         &                          & \begin{tabular}[c]{@{}c@{}}\textcolor{blue}{77.91}\\ \textcolor{red}{78.58}\end{tabular}                         &                          &                          & \begin{tabular}[c]{@{}c@{}}R: 1464\\ S: 1464\end{tabular}                         &                          & \begin{tabular}[c]{@{}c@{}}\textcolor{blue}{81.97}\\ \textcolor{red}{82.87}\end{tabular}                         \\
\multirow{-3}{*}{\begin{tabular}[c]{@{}l@{}}\textcolor{blue}{Merge with Real Data}\\ or \\ \textcolor{red}{Finetune on Real Data} \end{tabular}} & \cellcolor[HTML]{EFEFEF}$\checkmark$ & \cellcolor[HTML]{EFEFEF} & \cellcolor[HTML]{EFEFEF}$\checkmark$   & \cellcolor[HTML]{EFEFEF} & \cellcolor[HTML]{EFEFEF}$\checkmark$   & \cellcolor[HTML]{EFEFEF} & \cellcolor[HTML]{EFEFEF} & \cellcolor[HTML]{EFEFEF} & \cellcolor[HTML]{EFEFEF}\begin{tabular}[c]{@{}c@{}}R: 209\\ S: 418\end{tabular} & \cellcolor[HTML]{EFEFEF} & \cellcolor[HTML]{EFEFEF}\begin{tabular}[c]{@{}c@{}}\textcolor{blue}{78.52}\\ \textcolor{red}{80.21}\end{tabular} & \cellcolor[HTML]{EFEFEF} & \cellcolor[HTML]{EFEFEF} & \cellcolor[HTML]{EFEFEF}\begin{tabular}[c]{@{}c@{}}R: 1464\\ S: 2928\end{tabular} & \cellcolor[HTML]{EFEFEF} & \cellcolor[HTML]{EFEFEF}\begin{tabular}[c]{@{}c@{}}\textcolor{blue}{82.88}\\ \textcolor{red}{84.02}\end{tabular} \\ \hline
\end{tabular}%
}
\end{table}

\subsubsection{Data Ablation Study:}We evaluate the impact of synthetic data on semantic segmentation by comparing three strategies: (1) training on real data only, (2) training on synthetic data only, and (3) combining both. Results in Tab. \ref{tab:ab_data} show that while training solely on synthetic data achieves notable accuracy, it is lower than training on real data. However, combining synthetic with real data, either by merging or fine-tuning real data after pre-training on synthetic data, significantly improves performance. The best results are achieved when fine-tuning on real data after synthetic pre-training and using multiple synthetic datasets ($\mathcal{D}^{gen}_1$ and $\mathcal{D}^{gen}_2$) further enhances performance, demonstrating that synthetic data is effective when combined with real data. These results confirm that while synthetic data cannot fully replace real data, it plays a key role in improving model performance, especially when merged or fine-tuned with real data.

\subsubsection{Text prompt selection:}The performance of the model when selecting text prompts using different methods is detailed in Tab. \ref{tab:prompt}. Our proposed \textit{class-aware prompting} method demonstrates superior performance compared to previous methods. Specifically, our method achieves a performance of 78.52\%. These results indicate that our text prompt generation method helps the model focus more effectively on the classes that need to be segmented, thereby significantly improving the performance of the semantic segmentation model.

\subsubsection{Effect of different numbers of generated images in the synthetic data:}In addition to generating two synthetic images per original image (via Controllable Inpainting Diffusion and Img2Img Controllable Diffusion), we conducted experiments by increasing the number of generated images to evaluate semantic segmentation performance. The results in Table \ref{tab:number} are presented in the format \textit{X/Y}, where \textit{X} denotes the performance from merging synthetic data with real data, and \textit{Y} denotes the performance after fine-tuning on real data following pretraining. The results show that merging synthetic data with real data leads to degraded performance as the amount of synthetic data increases, whereas fine-tuning on real data after pretraining with synthetic data improves performance. These findings indicate that fine-tuning on real data after pretraining with a larger number of synthetic images can result in a more robust pre-trained model and improved overall performance.

\begin{table}[t]
\centering
\begin{minipage}{0.45\textwidth}
\centering
\caption{Performance of different text prompt selections when evaluating on VOC7 with Mask2Former (SwinB), The results are presented for model training using the merging mechanism.}
\label{tab:prompt}
\begin{tabular}{lllc}
\toprule
\textbf{Method}        &  &  & \textbf{mIoU (\%)} \\ \midrule
Simple text prompt     &  &  & 76.11       \\
Generated caption      &  &  & 75.67        \\
Class-prompt appending &  &  & 77.81        \\
Class-aware prompting  &  &  & 78.52        \\ \bottomrule
\end{tabular}%
\end{minipage}
\hfill
\begin{minipage}{0.52\textwidth}
\centering
\caption{Effect of increasing the number of synthetic images during training on the Mask2Former (SwinB) model. The values \textbf{N$_{real}$/N$_{syn}$} indicate the number of real and synthetic images, respectively.}
\label{tab:number}
\resizebox{1.04\textwidth}{!}{
\begin{tabular}{lclc}
\toprule
   \multicolumn{2}{c}{\textbf{VOC7}}               & \multicolumn{2}{c}{\textbf{VOC12}}              \\ \cmidrule(lr){1-2} \cmidrule(lr){3-4} 
   \textbf{N$_{real}$/N$_{syn}$} & \textbf{mIoU(\%)} & \textbf{N$_{real}$/N$_{syn}$} & \textbf{mIoU(\%)} \\ \midrule
209/0                     & 76.19       & 1464/0                   & 81.8       \\ \hline
 209/418                     & 78.52/80.21       & 1464/2928                   & 82.88/84.02       \\ \hline
 209/836                     & 79.21/81.53       & 1464/5856                   & 82.01/85.33       \\ \hline
 209/1254                    & 76.91/82.23       & 1464/8784                   & 80.08/87.05       \\ \bottomrule
\end{tabular}%
}
\end{minipage}
\end{table}

\section{ Discussion and Conclusion}

\subsection{Limitations}

Although our method shows promising results, there are several limitations. First, the quality of the synthesized images depends on the pre-trained generative model (SD). Second, generating high-quality synthetic images using diffusion models can be computationally expensive and time-consuming. Finally, while our method has potential for privacy-sensitive applications, this study only evaluates general datasets, so further validation is needed to ensure its effectiveness in specific scenarios.

\subsection{Conclusion}

In this work, we proposed a novel synthetic data augmentation pipeline that combines controllable diffusion models with advanced conditioning techniques to tackle the challenges of balancing diversity and reliability in semantic segmentation. Our method effectively generates high-quality synthetic data that preserves the structure of labeled objects and aligns well with real-world data distributions, demonstrating significant performance improvements on benchmark datasets like PASCAL VOC and BDD100K, particularly in data-scarce scenarios. Moreover, our approach effectively mitigates domain shift issues commonly associated with synthetic data generation, enabling more robust training.

Building on the success of image transformations guided by segmentation masks, we explore their potential for privacy protection applications. In privacy scenarios, sensitive regions identified by segmentation masks can be concealed using techniques like inpainting, ensuring privacy while preserving the quality of synthetic datasets for training segmentation models. Further exploration of these methods could enhance their applicability in privacy-sensitive domains.

\section{Acknowledgement}

This research is funded by Vietnam National University Ho Chi Minh City (VNU-HCM) under grant C2023-26-10.

{\small
\bibliographystyle{IEEEtran}
\bibliography{ref}
}

\end{document}